\documentclass[11pt]{article}

\usepackage[preprint]{acl}

\usepackage{times}
\usepackage{latexsym}
\usepackage{amsmath}
\usepackage{amssymb}
\usepackage[T1]{fontenc}
\usepackage{booktabs}
\usepackage{multirow}
\usepackage[utf8]{inputenc}

\usepackage{microtype}

\usepackage{inconsolata}

\usepackage{graphicx}

\usepackage{placeins}

\title{Prompt Embedding Probes (PEP):\\Hallucination Detection in LLMs from Hidden States}

\author{Zakhar Mrykhin \\
  \texttt{zazamrykh@gmail.com} \\\And
  Valentin Malykh \\
  HSE University\\
  \texttt{valentin.malykh@phystech.edu} \\}

\begin{document}
\maketitle
\begin{abstract}
Large language models (LLMs) can generate fluent and useful responses, but they remain prone to hallucinations. This motivates efficient methods for detecting unreliable outputs at inference time. In this paper, we study answer-level hallucination detection in a white-box setting based on hidden states of a frozen LLM, and introduce Prompt Embedding Probes, a parameter-efficient extension of standard linear probes that augments the input with a small number of learnable prompt embeddings.

We evaluate PEP on TriviaQA, GSM8K, and MedQA using Qwen3 models of multiple scales. The results show that learnable prompt embeddings improve hidden-state-based hallucination detection in the main in-distribution setting relative to standard linear probes. We further study the method in a to-be-generated (TBG) setting for pre-generation hallucination prediction, in cross-model transfer experiments, and under out-of-distribution evaluation. PEP remains effective in the TBG and cross-model settings, while robust transfer across substantially different datasets remains difficult.

These results suggest that prompt-based adaptation can strengthen probe-based hallucination detection while keeping the backbone frozen and adding only a small number of trainable parameters.
\end{abstract}

% \section{Introduction}

% Large language models (LLMs) are capable of producing fluent and useful responses, yet they still remain prone to hallucinations—outputs that are incorrect, unsupported, or otherwise unreliable \citep{huang2025survey}.

% This is especially concerning in high-risk domains such as medicine, finance, and law, where model errors may lead to consequential decisions or misleading professional advice. Legal cases already show that hallucinated content in LLM-assisted workflows can result in tangible penalties and procedural harm \citep{reuters_ai_lawyer_2026}. 

% Moreover, hallucinations remain common in widely used evaluation datasets and benchmarks indicating that they persist despite continued progress in LLM development \citep{niu-etal-2024-ragtruth, li-etal-2023-halueval}. 

% Prior work suggests that hallucinations arise from multiple sources spanning the data, training, and inference stages, which makes them difficult to eliminate with a single intervention \citep{huang2025survey}. Previous work also provides illustrative examples of hallucinations and discusses why they arise, showing that even simple factual prompts can elicit confident but incorrect answers from LLMs \citep{azaria-mitchell-2023-internal}. 

% This motivates hallucination detection as a complementary direction: alongside efforts to reduce hallucinations, one may also seek to identify unreliable model outputs at inference time.

\section{Introduction}

Large language models (LLMs) are capable of producing fluent and useful responses, yet they remain prone to hallucinations---outputs that are incorrect, unsupported, or otherwise unreliable \citep{huang2025survey}. This is especially concerning in high-risk domains such as medicine, finance, and law, where model errors may lead to consequential decisions or misleading professional advice. Legal cases already show that hallucinated content in LLM-assisted workflows can result in tangible penalties and procedural harm \citep{reuters_ai_lawyer_2026}. Hallucinations also remain common in widely used evaluation datasets and benchmarks, indicating that they persist despite continued progress in LLM development \citep{niu-etal-2024-ragtruth,li-etal-2023-halueval}.

Hallucinations can arise from multiple sources across the data, training, and inference stages, making them difficult to eliminate through a single intervention \citep{huang2025survey}. Prior work further shows that even simple factual prompts can elicit confident but incorrect answers from LLMs \citep{azaria-mitchell-2023-internal}. Hallucination detection therefore complements efforts to reduce hallucinations by identifying unreliable model outputs at inference time.

Among existing approaches, white-box methods are particularly appealing because some of them are simple and efficient at inference time. A growing body of work suggests that hidden states and related internal representations contain useful signals for detecting hallucinations and assessing answer reliability \citep{azaria-mitchell-2023-internal, kossen2024semantic, snyder2024early, zhang2025detecting, chen2025persona, obeso2025real, han-etal-2025-simple}.
At the same time, standard linear probes may be limited in how flexibly they adapt to the hallucination detection task when applied directly to frozen hidden states.

This paper investigates whether hidden-state-based hallucination detection can be improved by augmenting probes with learnable prompt embeddings \citep{lester-etal-2021-power}. To this end, we introduce Prompt Embedding Probes, a white-box method that extends standard linear probes with trainable prompt embeddings for the detection of hallucinations at answer-level in LLMs. The experiments evaluate PEP on TriviaQA, GSM8K, and MedQA using Qwen3 models at multiple scales, and compare it against linear probes and additional baselines. 

The main contributions of this paper are as follows:
\begin{itemize}
    \item We introduce Prompt Embedding Probes, a white-box method for answer-level hallucination detection in LLMs based on hidden states and learnable prompt embeddings. The code used in this study is available at \href{https://github.com/zazamrykh/internal_probing}{\texttt{github.com/zazamrykh/internal\_probing}}.
    \item We show that learnable prompt embeddings improve in-distribution hallucination detection relative to standard linear probes across multiple datasets and model scales.
    \item We provide an empirical analysis of PEP beyond the main in-distribution setting, including out-of-distribution evaluation, cross-model transfer, and a to-be-generated token setting for pre-generation hallucination prediction. We also study the effect of key method parameters such as prompt embedding placement, prompt embedding count, and probing layer selection.
\end{itemize}

\section{Related Work}

Hallucination detection in LLMs has been studied in different settings. Black-box methods typically rely on response consistency in repeated sampling \citep{manakul-etal-2023-selfcheckgpt}, self-verification or verification-based checks \citep{dhuliawala-etal-2024-chain}, or external verification pipelines \citep{min-etal-2023-factscore}. Such approaches can be effective, but they are often computationally expensive and may require multiple model calls or additional verification components.

Other approaches use confidence signals: token probabilities or related log-probability statistics.
Examples include probability-based self-evaluation methods such as \textit{P(True)} \citep{kadavath2022language} and token-level uncertainty methods for fact-checking generated claims \citep{fadeeva-etal-2024-fact}. 
However, token-level uncertainty is inherently limited: low-probability tokens may reflect not only uncertainty about the underlying meaning, but also lexical and syntactic variation \citep{farquhar2024detecting}.

White-box approaches to hallucination detection use internal LLM representations. \citet{azaria-mitchell-2023-internal}, \citet{snyder2024early}, \citet{chen2025persona}, and related work showed that hidden states and other generation-time artifacts contain useful signals for hallucination detection. \citet{kossen2024semantic} demonstrated that simple linear probes on hidden states are effective for predicting targets such as answer correctness and semantic entropy. \citet{zhang2025detecting} proposed a method that selects informative neurons from internal representations and uses them for hallucination detection. \citet{han-etal-2025-simple} used linear probes for factual hallucination detection at the claim level, which is useful because LLM responses often contain multiple claims that require verification. \citet{obeso2025real} studied the applicability of linear probes to real-time factual hallucination detection at the entity level. They also explored LoRA adapters to improve detection quality while using regularization to preserve the model's generation behavior.

Overall, prior work suggests that linear probes are a practical and effective family of white-box methods for hallucination detection across different targets and granularities. At the same time, existing work mainly studies which internal representations or probing targets are most useful and focuses on different hallucination types, while the effect of modifying the probing input through learnable prompt embeddings remains underexplored. This paper addresses that gap by studying whether trainable prompt embeddings can improve hidden-state-based hallucination detection over standard linear probes.

\section{Method}

\subsection{Task Formulation}

This work considers answer-level hallucination detection. Given a query \(x\) and a model-generated answer \(a\), the goal is to predict a binary label \(y \in \{0,1\}\), where \(y=1\) denotes a hallucinated answer. More generally, a detector defines a scoring function
\[
f_\theta(x,a) \rightarrow [0,1].
\]
In the to-be-generated (TBG) setting, detection is performed before the answer is generated, while in the second-last-token (SLT) setting it is performed using representations obtained after the answer has been processed.

\subsection{Linear Probe Baseline}

Let \(z=(z_1,\dots,z_T)\) be the token sequence given to the model, and let
\[
H^{(\ell)} = (h_1^{(\ell)}, \dots, h_T^{(\ell)}), \qquad h_t^{(\ell)} \in \mathbb{R}^d,
\]
denote the hidden states after layer \(\ell\) of a frozen LLM. The linear probe uses a hidden state selected by the probing layer \(\ell\) and probing position \(q\):
\[
u = h_q^{(\ell)}.
\]
A linear classification head is then applied on top of \(u\):
\[
\hat{p}_\theta(y=1 \mid x,a) = \sigma(w^\top u),
\]
where \(w \in \mathbb{R}^d\), \(\sigma(\cdot)\) is the sigmoid function, and \(\theta = w\) are the trainable parameters of the detector.

\subsection{Prompt Embedding Probe}

PEP extends the linear probe baseline by inserting \(M\) trainable prompt embeddings into the input embedding sequence before the forward pass through the frozen LLM. Let
\[
E(z) = (e_1,\dots,e_T), \qquad e_t \in \mathbb{R}^d,
\]
be the input embeddings of \(z\), and let
\[
P = (p^{(1)}, \dots, p^{(M)}), \qquad p^{(j)} \in \mathbb{R}^d,
\]
be the trainable prompt embeddings. After insertion, we obtain an augmented sequence
\[
\tilde{E} = I(E(z), P),
\]
where \(I(\cdot)\) denotes the chosen insertion scheme.

Let
\[
\tilde{H}^{(\ell)} = (\tilde{h}_1^{(\ell)}, \dots, \tilde{h}_{T+M}^{(\ell)})
\]
be the hidden states after layer \(\ell\) produced from the modified input. As in the baseline, detection is performed at probing layer \(\ell\) and probing position \(q\), where \(q\) is defined with respect to the original token sequence before prompt insertion. Denoting by \(\pi(q)\) the corresponding position after insertion, the detector uses
\[
u = \tilde{h}_{\pi(q)}^{(\ell)}.
\]
A linear head is applied on top of the selected hidden state:
\[
\hat{p}_\theta(y=1 \mid x,a) = \sigma(w^\top u),
\]
where
\[
\theta = \left(\{p^{(j)}\}_{j=1}^{M}, w\right)
\]
is the set of trainable parameters. Thus, unlike the baseline, PEP jointly learns both the probe weights and the inserted prompt embeddings, while keeping the base LLM frozen. 

PEP adds only the prompt embeddings and the classification head as trainable parameters. For $M$ prompt embeddings and hidden size $d$, the prompt component contains $M \times d$ trainable parameters, while the linear head contains $d+1$ parameters when implemented with a bias. Thus, PEP trains $M \times d + d + 1$ parameters in total, while all backbone parameters remain frozen.

\begin{figure}[t]
    \centering
    \includegraphics[width=\columnwidth]{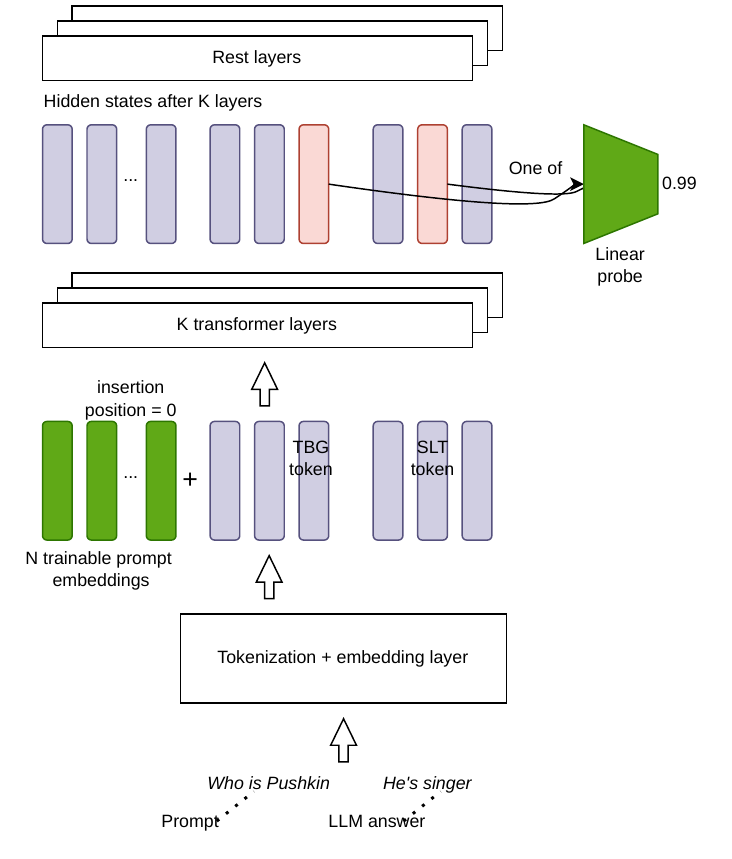}
    \caption{Overview of Prompt Embedding Probe. Trainable prompt embeddings are inserted into the input embedding sequence, processed by a frozen LLM, and the selected hidden state is passed to a linear classification head for hallucination detection.}
    \label{fig:pep_overview}
\end{figure}

Both methods are trained with binary cross-entropy on answer-level labels. In practice, the linear probe is implemented with \texttt{LogisticRegression} from \texttt{scikit-learn} using \texttt{lbfgs} and \(C=1.0\), while PEP is implemented in \texttt{PyTorch} with Hugging Face \texttt{transformers}.

\section{Experimental Setup}

We evaluate on TriviaQA \citep{joshi-etal-2017-triviaqa}, GSM8K \citep{cobbe2021training}, and MedQA \citep{jin2021disease}. TriviaQA targets short factual question answering, GSM8K focuses on mathematical reasoning, and MedQA represents a high-risk medical QA setting with multiple-choice answers. For TriviaQA, we use the system prompt \textit{``Answer briefly.''} For GSM8K, we use the system prompt \textit{``Solve the problem step by step. At the end write exactly: Final answer: <number>''} The exact prompt used for MedQA is provided in Appendix~\ref{app:medqa-prompt}.

\textbf{Baselines.} We compare PEP with three baselines. First, the \textbf{linear probe} applies a logistic-regression classifier to a selected hidden state of the frozen LLM. Second, \textbf{average negative log-probability} uses the average negative log-probability of the generated answer as an uncertainty-based score. Third, \textbf{P(True) single} is a single-generation self-evaluation baseline: the model is prompted once to assess whether its generated answer is true or false, and the resulting probability is used as the detection score.

\begin{table}[t]
\centering
\small
\begin{tabular}{|l|c|c|c|}
\hline
Dataset & Train & Valid & Test \\
\hline
TriviaQA & 10,000 & 1,000 & 17,944 \\
GSM8K & 6,473 & 1,000 & 1,319 \\
MedQA & 8,000 & 1,000 & 1,273 \\
\hline
\end{tabular}
\caption{Dataset splits used in the experiments.}
\label{tab:datasets}
\end{table}

\begin{figure*}[t]
    \centering
    \includegraphics[width=\textwidth]{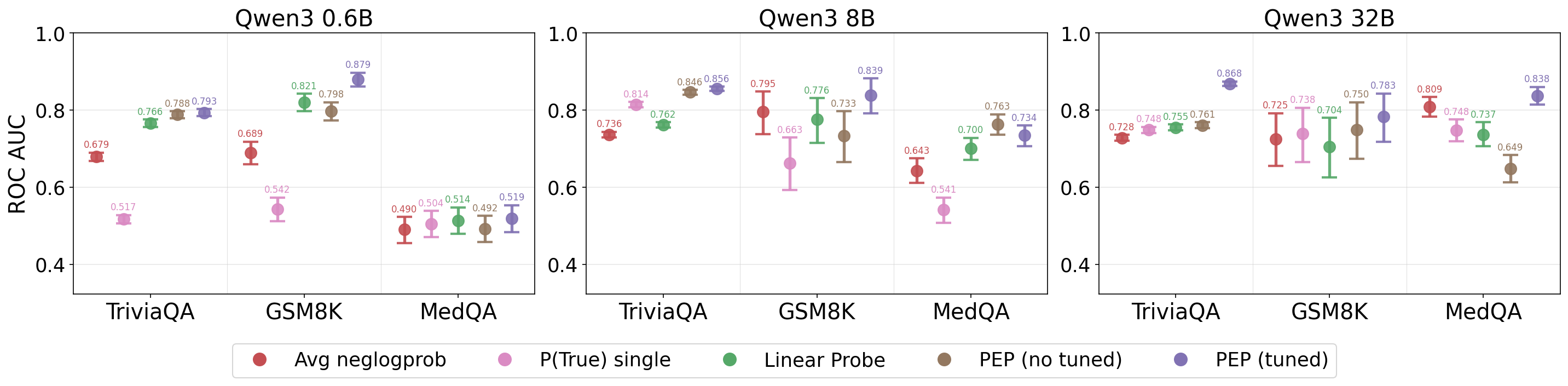}
    \caption{ROC AUC with bootstrap 95\% confidence intervals for PEP comparison with baselines across three datasets and three Qwen3 model sizes.}
    \label{fig:main_results}
\end{figure*}

\begin{figure*}[t]
    \centering
    \includegraphics[width=\textwidth]{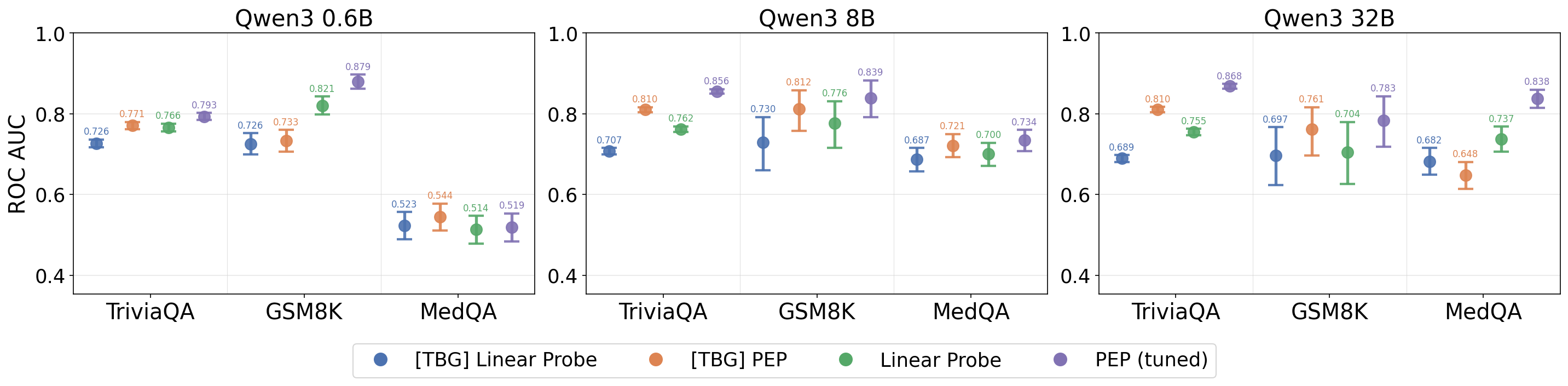}
    \caption{ROC AUC with bootstrap 95\% confidence intervals for PEP vs linear probe in TBG and SLT setups}
    \label{fig:tbg_results}
\end{figure*}

For TriviaQA, correctness is evaluated with substring match under a short-answer prompting format. For GSM8K and MedQA, correctness is determined with format-based matching rules.

We report ROC AUC in all experiments. 

We use Qwen3 models \citep{yang2025qwen3} in three sizes: 0.6B, 8B, and 32B parameters. This allows us to evaluate the method across substantially different model scales within the same model family. For cross-model and OOD experiments we also use Gemma3 12B \citep{gemma3}.

Experiments were conducted on an NVIDIA H200 GPU with 140 GB of memory using PyTorch and Hugging Face Transformers. As a representative example, training PEP with 5 prompt embeddings on Qwen3 8B for TriviaQA required approximately 85 GPU minutes, estimated from training logs up to convergence. In the more expensive setting of GSM8K with Qwen3 32B, we used batch size 1 and observed a training time of approximately 27 hours up to the best validation ROC AUC. The exact training configuration and implementation details are available in the public repository.

\section{Results}

\begin{table*}[t]
\centering
\small
\caption{
Main in-distribution results. ROC AUC of the linear probe and tuned PEP across three datasets and Qwen3 model sizes. $\Delta$ ROC AUC is the paired-bootstrap estimate of the difference between tuned PEP and the linear probe; positive values favor PEP. For Qwen3-0.6B on MedQA, the confidence interval includes zero, so we do not interpret the observed difference as evidence of a reliable improvement.
}

\label{tab:pep-vs-probe-diff}
\begin{tabular}{llcccc}
\toprule
Model & Dataset & Linear Probe & Tuned PEP & $\Delta$ ROC AUC & 95\% CI \\
\midrule
\multirow{3}{*}{Qwen3-0.6B}
& TriviaQA & 0.766 & 0.793 & +0.027 & [0.020, 0.034] \\
& GSM8K    & 0.821 & 0.879 & +0.059 & [0.040, 0.077] \\
& MedQA    & 0.514 & 0.519 & +0.005 & [-0.037, 0.048] \\
\midrule
\multirow{3}{*}{Qwen3-8B}
& TriviaQA & 0.762 & 0.856 & +0.094 & [0.088, 0.100] \\
& GSM8K    & 0.776 & 0.839 & +0.062 & [0.011, 0.117] \\
& MedQA    & 0.700 & 0.734 & +0.034 & [0.002, 0.065] \\
\midrule
\multirow{3}{*}{Qwen3-32B}
& TriviaQA & 0.755 & 0.868 & +0.113 & [0.106, 0.121] \\
& GSM8K    & 0.704 & 0.783 & +0.079 & [0.012, 0.146] \\
& MedQA    & 0.737 & 0.838 & +0.101 & [0.071, 0.130] \\
\bottomrule
\end{tabular}
\end{table*}

\subsection{Prompt embeddings improve hallucination detection}

We begin with the main question of this work: do prompt embeddings improve hidden-state-based hallucination detection? Figure~\ref{fig:main_results} summarizes the main comparison.

Figure~\ref{fig:main_results} compares tuned and untuned PEP with the linear-probe and probability-based baselines described in Section~4. It also reports PEP with fixed untuned hyperparameters and tuned PEP. For the linear probe, the best probing layer is selected on the validation set. In contrast, the untuned PEP variant uses a fixed configuration chosen a priori before running the experiments: probing at 0.6 of the total model depth, 10 prompt embeddings inserted at the beginning of the sequence, and the SLT setup.

The main result is that tuned PEP outperforms the linear probe in 8 of 9 model--dataset combinations under the paired-bootstrap comparison. The only exception is Qwen3-0.6B on MedQA, where the estimated improvement is small and its confidence interval includes zero. We therefore do not interpret this comparison as evidence of a reliable improvement. Table~\ref{tab:pep-vs-probe-diff} reports the absolute ROC AUC values together with the paired-bootstrap differences and confidence intervals.

\subsection{PEP remains effective in the TBG setting}

We next test whether the gains from prompt embeddings persist in the more challenging to-be-generated (TBG) setting. Using the configurations selected in the SLT experiments, we evaluate PEP before answer generation is complete and compare it against the linear probe baseline. Figure~\ref{fig:tbg_results} shows the results of this experiment.

The result is encouraging: TBG PEP outperforms even the SLT linear probe in 8 of 9 model--dataset combinations under a one-sided paired bootstrap test on the ROC AUC difference with significance level 0.05. Overall, these results show that the benefit of prompt embeddings is not restricted to the SLT setup and extends to the more difficult TBG regime.

\subsection{Cross-model transfer}

\begin{figure}[t]
    \centering
    \includegraphics[width=\columnwidth]{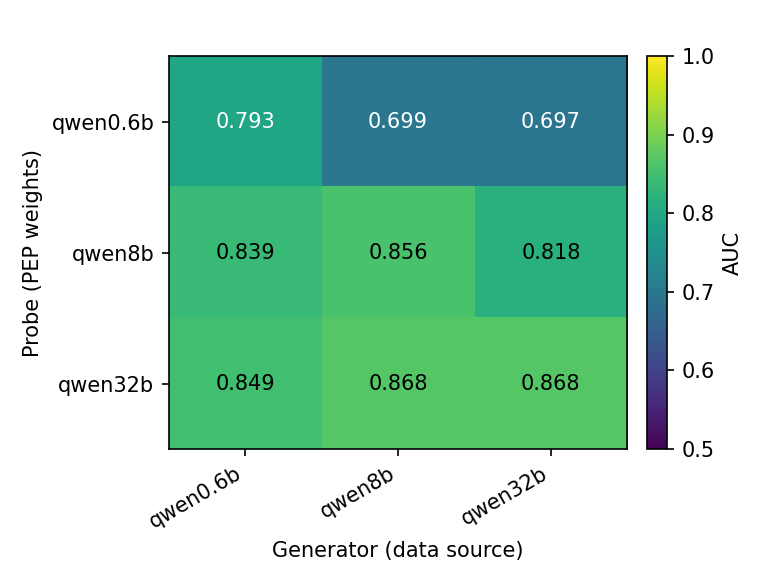}
    \caption{Cross-model transfer within the Qwen3 family on TriviaQA. Each detector is trained on its own model's generations and evaluated on generations produced by all models in the comparison.}
    \label{fig:cross_models_qwen}
\end{figure}

\begin{figure}[t]
    \centering
    \includegraphics[width=\columnwidth]{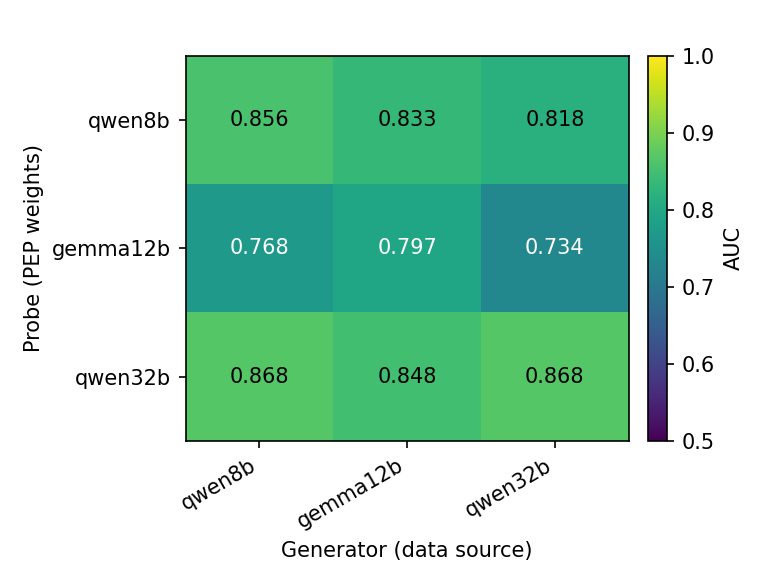}
    \caption{Cross-model transfer across model families on TriviaQA. Gemma 3 12B is used without additional parameter tuning.}
    \label{fig:cross_models_gemma}
\end{figure}

We next study whether a detector based on one model can identify hallucinations in answers generated by another model. This setting separates the generator from the detector and therefore moves the evaluation toward a black-box-like regime with respect to the target generator, while still allowing white-box access to the detector model itself.

All experiments in this section are conducted on TriviaQA. Importantly, each detector is trained on the generations of its own base model and is then evaluated both on its own generations and on generations produced by other models. We consider two transfer settings. In the first, we compare Qwen3 0.6B, 8B, and 32B in order to study transfer across model scales within the same family (see Figure ~\ref{fig:cross_models_qwen}). In the second, we compare Qwen3 8B, Qwen3 32B, and Gemma 3 12B in order to test transfer across model families (see Figure ~\ref{fig:cross_models_gemma}). Gemma 3 12B is included without additional parameter tuning.

The results can be read from two complementary perspectives. First, row-wise, each detector generally performs best on the generations of its own model. This suggests that model-specific generation patterns are indeed useful for hallucination detection. Second, column-wise, the best detector for a given generator is typically the largest one, namely Qwen3 32B. Thus, for detection quality, detector scale appears to matter more than using the hidden states of exactly the same model that produced the answer.

Overall, the experiment leads to three conclusions. First, hidden-state-based hallucination detection can be applied to answers generated by a different model, including models from another family. Second, smaller detector models generally transfer worse to stronger generators, although their performance remains reasonably competitive. Third, detector size plays an important role in final performance, suggesting that exact generator--detector identity is not the dominant factor in cross-model detection.

\subsection{Effect of probing layer}

\begin{figure}[t]
    \centering
    \includegraphics[width=\columnwidth]{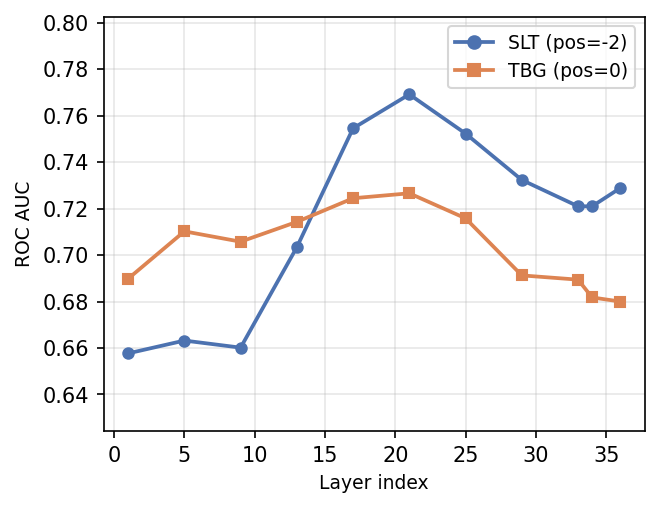}
    \caption{ROC AUC as a function of probing layer for Qwen3 8B on TriviaQA in the SLT and TBG settings.}
    \label{fig:qwen8b_triviaqa_layer_pos}
\end{figure}

We next study how the choice of probing layer affects hallucination detection quality. Figure~\ref{fig:qwen8b_triviaqa_layer_pos} shows one representative example, while the full set of 3$\times$3 plots for all model--dataset combinations is provided in Appendix~\ref{app:full-plots}.

The results show that the optimal layer depends on the evaluation setup, the model, and the dataset. Typically the best probing layer lies between the middle and the end of the network rather than at a fixed relative depth.

A more stable way to describe the best probing position is by its distance to the final layer rather than by the fraction of total model depth. Table~\ref{tab:best-layer-by-model} summarizes this pattern across models. Table~\ref{tab:appendix-best-layer-by-dataset} describes the median best fraction of total depth and the number of layers to the end for the best layer choice.

\begin{table}[t]
\centering
\small
\setlength{\tabcolsep}{4pt}
\begin{tabular}{|l|c|c|c|}
\hline
 & Qwen3 0.6B (28) & Qwen3 8B (36) & Qwen3 32B (64) \\
\hline
TBG & 19 / 0.68 / 9 & 25 / 0.69 / 11 & 53 / 0.83 / 11 \\
SLT & 19 / 0.68 / 9 & 25 / 0.69 / 11 & 57 / 0.89 / 7 \\
\hline
\end{tabular}
\caption{Median best probing layer across datasets, reported as layer index / fraction of total depth / number of layers to the end. Headers contain maximum number of layers for each model.}
\label{tab:best-layer-by-model}
\end{table}

\begin{table}[t]
\centering
\small
\setlength{\tabcolsep}{4pt}
\begin{tabular}{|l|c|c|c|}
\hline
 & TriviaQA & GSM8K & MedQA \\
\hline
TBG & 0.58 / 15 & 0.69 / 11 & 0.83 / 9 \\
SLT & 0.68 / 15 & 0.69 / 9 & 0.89 / 3 \\
\hline
\end{tabular}
\caption{Median best probing layer aggregated by dataset and reported as fraction of total depth / number of layers to the end.}
\label{tab:appendix-best-layer-by-dataset}
\end{table}

\subsection{Effect of insertion position and number of prompt embeddings}

We next study how the method depends on two additional hyperparameters: the insertion position of the prompt embeddings and their number. For the probing layer selected in the SLT setup, we sweep three insertion strategies---beginning only, end only, and both beginning and end---as well as the number of inserted embeddings. Table ~\ref{tab:embedding-search-summary} summarizes results of this experiment. Full results for all model--dataset combinations are provided in Appendix~\ref{app:full-plots}.

The best configuration depends on both the model and the dataset. But at the same time in most experiments, inserting a small number of prompt embeddings at the beginning of the sequence gives the best result, while insertion only at the end wins only once.

\begin{table}[t]
\centering
\small
\setlength{\tabcolsep}{6pt}
\begin{tabular}{|l|c|c|}
\hline
Insertion strategy & Wins (out of 9) & Median best count \\
\hline
Beginning only & 6 & 5 \\
End only & 1 & 3 \\
Both & 2 & 3 \\
\hline
\end{tabular}
\caption{Aggregate results of the embedding search across the nine model--dataset configurations.}
\label{tab:embedding-search-summary}
\end{table}

\subsection{Out-of-distribution evaluation}

We study an out-of-distribution (OOD) setting in which the detector is trained on two of the three datasets and evaluated on the held-out third one. Both the generator and detector LLM were the same in this experiment: Gemma3 12B. Results are available in Table ~\ref{tab:ood-results}.

\begin{table}[t]
\centering
\small
\begin{tabular}{|l|c|c|}
\hline
Held-out dataset & Linear probe & PEP \\
\hline
TriviaQA & 0.5490 & 0.6250 \\
MedQA & 0.5117 & 0.5267 \\
GSM8K & 0.5676 & 0.5163 \\
\hline
\end{tabular}
\caption{OOD evaluation: training on two datasets and testing on the held-out third one.}
\label{tab:ood-results}
\end{table}

Overall, both PEP and the linear probe perform poorly in this setting. A likely reason is that the three datasets differ substantially in both the answer format and error type, making cross-dataset transfer difficult for both methods.

The only clearly meaningful result is obtained when TriviaQA is used as the held-out dataset. But overall the gains of PEP are most convincing in the in-distribution setting, while robust OOD transfer remains an open challenge. 

\section{Conclusion}

We introduced Prompt Embedding Probes (PEP), a white-box method for answer-level hallucination detection based on hidden states of a frozen LLM. PEP extends standard linear probes by jointly learning a small number of prompt embeddings and a linear classification head, while keeping all backbone parameters frozen.

Across three QA datasets and multiple Qwen3 model sizes, PEP improves in-distribution ROC AUC relative to a standard linear probe in most evaluated settings. The gains also extend to the to-be-generated setting and to the cross-model experiments considered in this work. However, the cross-dataset OOD results show that neither PEP nor the linear probe transfers robustly across substantially different datasets. Thus, the current evidence supports PEP primarily as an in-distribution prompt-based adaptation for hidden-state probing, rather than as a universally transferable hallucination detector.

Overall, PEP provides a simple parameter-efficient extension of linear probing, although its training is more computationally demanding than fitting a linear probe on cached hidden states. Improving cross-dataset generalization and comparing prompt-based adaptation with stronger probe architectures and alternative adaptation methods remain important directions for future work.

\section{Potential Risks}

Although the proposed method is intended to improve reliability monitoring, incorrect detector predictions may still be harmful in practice. False negatives may lead users to trust unreliable model outputs, while false positives may suppress correct answers and reduce system utility. In addition, our results show limited robustness under cross-dataset transfer, so deploying the detector outside the evaluated settings may create a misleading sense of safety, especially in high-stakes domains such as medicine. For this reason, we view the proposed method as a supporting signal for risk-aware use rather than as a standalone guarantee of answer reliability.

\section{Artifacts, Licensing, Intended Use}
This work uses publicly available datasets and model families for research and evaluation purposes, consistent with their original intended use and respective terms of use. The implementation associated with this work is publicly available under the MIT license at \href{https://github.com/zazamrykh/internal_probing}{\texttt{github.com/zazamrykh/internal\_probing}} and is intended for research use. We do not claim that the resulting detector is ready for deployment in safety-critical settings without additional validation.

\section*{Limitations}

Despite its empirical improvements over standard linear probes, the proposed method has several limitations. First, Prompt Embedding Probes are more complex both architecturally and computationally than standard hidden-state-based linear probing. The method requires a more involved implementation and more expensive training, since optimizing prompt embeddings requires repeated forward passes through the frozen language model together with gradient computation, which increases GPU memory usage and training time.

Second, the OOD experiments show that the method does not transfer well across substantially different datasets. This likely reflects large differences in answer format and task structure between the considered benchmarks, suggesting that stronger cross-dataset generalization remains an open problem. One possible direction for future work is to explore alternative training targets, such as uncertainty- or semantic entropy-related signals.

Third, the current study focuses on answer-level hallucination detection, whereas many practical applications increasingly require claim-level detection for long-form responses. Although the proposed method could likely be adapted to finer-grained prediction, this was not investigated in the present work.

Finally, the method depends on several hyperparameters, including the probing layer, insertion position, and number of prompt embeddings. Even though PEP shows strong performance under non-optimal settings, obtaining the highest detection quality still requires tuning these parameters for a particular setup.

\bibliography{anthology,custom}
% \bibliography{custom}

% \section{Example Appendix}
% \label{sec:appendix}
\FloatBarrier
\FloatBarrier

\appendix

\section{MedQA prompt}
\label{app:medqa-prompt}

This appendix provides the exact system prompt used for MedQA. In contrast to TriviaQA and GSM8K, MedQA requires a strict answer format because correctness is determined from the selected option letter.

\begin{quote}
\small
You are answering a medical multiple-choice question.\\
Choose the single best answer from the provided options.\\
Output exactly in the format:\\
Answer: <LETTER>\\
Do not provide any explanation.
\end{quote}

\section{Full plots}
\label{app:full-plots}

\begin{figure*}[t]
    \centering
    
    \textbf{Additional probing-layer results}

    \vspace{0.3em}
    \includegraphics[width=\textwidth,height=0.5\textheight,keepaspectratio]{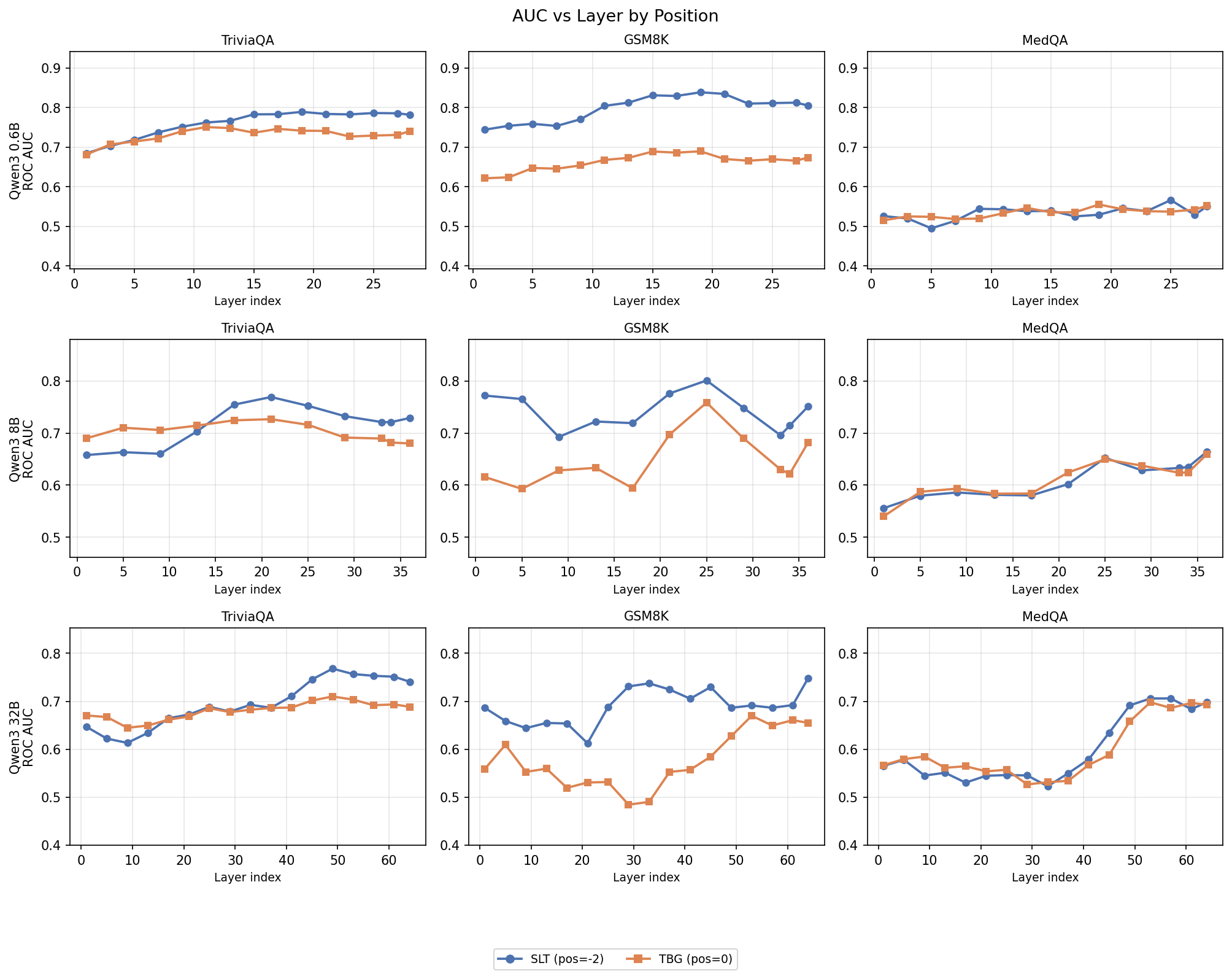}

    \vspace{0.8em}
    \textbf{Additional embedding-search results}

    \vspace{0.3em}
    \includegraphics[width=\textwidth,height=0.4\textheight,keepaspectratio]{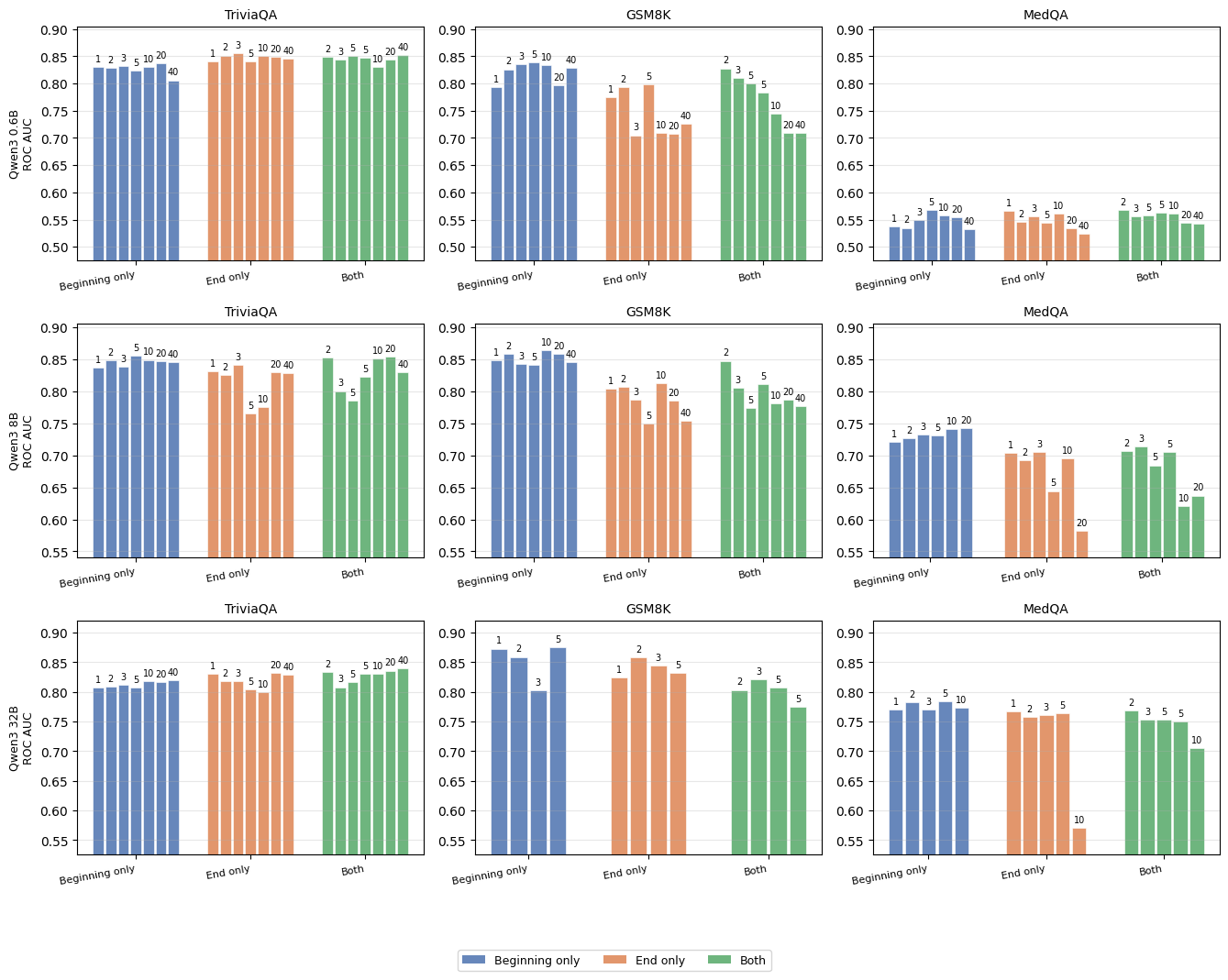}

    \caption{Full probing-layer sweeps for all three models and three datasets and full embedding-search results for all model--dataset combinations.}
    \label{fig:appendix-results}
\end{figure*}

This appendix provides the full plots for the two parameter studies discussed in the main text. Figure~\ref{fig:appendix-results} includes the complete probing-layer sweeps for all model--dataset combinations and the full embedding-search results over insertion strategies and the number of prompt embeddings.

\end{document}